\documentclass[letterpaper, 10 pt, conference]{ieeeconf}  

\IEEEoverridecommandlockouts                              

\usepackage{graphics} 
\usepackage{epsfig} 
\usepackage{mathptmx} 
\usepackage{times} 
\usepackage{amsmath} 
\usepackage{amssymb}  
\usepackage{multirow}
\usepackage{booktabs}
\usepackage{algorithm}
\usepackage{algorithmic}

\newcommand*\squeezespaces[1]{
  \thickmuskip=\scalemuskip{\thickmuskip}{#1}%
  \medmuskip=\scalemuskip{\medmuskip}{#1}%
  \thinmuskip=\scalemuskip{\thinmuskip}{#1}%
  \nulldelimiterspace=#1\nulldelimiterspace
  \scriptspace=#1\scriptspace
}
\newcommand*\scalemuskip[2]{%
  \muexpr #1*\numexpr\dimexpr#2pt\relax\relax/65536\relax
} 

\title{\LARGE \bf
Introspective Visuomotor Control: Exploiting Uncertainty in Deep Visuomotor Control for Failure Recovery
}

\author{Chia-Man Hung$^{1,2}$, Li Sun$^{1}$, Yizhe Wu$^{1}$, Ioannis Havoutis$^{2}$, Ingmar Posner$^{1}$
\thanks{$^{1}$Applied AI Lab (A2I), $^{2}$Dynamic Robot Systems (DRS)}%
\thanks{Oxford Robotics Institute (ORI), University of Oxford}
\thanks{Correspondence to: {\tt\small chiaman@robots.ox.ac.uk}}%
}

\begin{document}

\maketitle
\thispagestyle{empty}
\pagestyle{empty}

\begin{abstract}

End-to-end visuomotor control is emerging as a compelling solution for robot manipulation tasks.
However, imitation learning-based visuomotor control approaches tend to suffer from a common limitation, lacking the ability to recover from an out-of-distribution state caused by compounding errors.
In this paper, instead of using tactile feedback or explicitly detecting the failure through vision, we investigate using the uncertainty of a policy neural network.
We propose a novel uncertainty-based approach to detect and recover from failure cases.
Our hypothesis is that policy uncertainties can implicitly indicate the potential failures in the visuomotor control task and that robot states with minimum uncertainty are more likely to lead to task success.
To recover from high uncertainty cases, the robot monitors its uncertainty along a trajectory and explores possible actions in the state-action space to bring itself to a more certain state.
Our experiments verify this hypothesis and show a significant improvement on task success rate: 12\% in pushing, 15\% in pick-and-reach and 22\% in pick-and-place.

\end{abstract}

\section{INTRODUCTION}

Deep visuomotor control (VMC) is an emerging research area for closed-loop robot manipulation, with applications in dexterous manipulation, such as manufacturing and packing. Compared to conventional vision-based manipulation approaches, deep VMC aims to learn an end-to-end policy to bridge the gap between robot perception and control, as an alternative to explicitly modelling the object position/pose and planning the trajectories in Cartesian space.

The existing works on deep VMC mainly focus on domain randomisation~\cite{james2017transferring}, to transfer visuomotor skills from simulation to the real world \cite{bousmalis2018using,james2019sim}; or one-shot learning~\cite{duan2017one, finn2017one}, to generalise visuomotor skills to novel tasks when large-scale demonstration is not available. In these works, imitation learning is used to train a policy network to predict motor commands or end-effector actions from raw image observations. Consequently, continuous motor commands can be generated, closing the loop of perception and manipulation. However, with imitation learning, the robot may fall into an unknown state-space to which the policy does not generalise, where it is likely to fail.
Early diagnosis of failure cases is thus important for policy generalisation but an open question in deep VMC research.


Instead of using vision or tactile feedback to detect failure cases~\cite{kragic2002visually,sun2017single}, we extend the widely-used deterministic policy network to an introspective Bayesian network. The uncertainty obtained by this Bayesian network is then used to detect the failure status. More importantly, as a supplement to the existing deep VMC methods, we propose a recovery mechanism to rescue the manipulator when a potential failure is detected, where a predictive model can learn the intuitive uncertainty to indicate the status of manipulation without the need of simulating the manipulation using a physics engine.


In summary, our contributions are three-fold:
First, we extend VMC to a probabilistic model which is able to estimate its epistemic uncertainty.
Second, we propose a simple model to predict the VMC policy uncertainty conditioned on the action without simulating it.
Finally, leveraging the estimated policy uncertainty, we propose a strategy to detect and recover from failures, thereby improving the success rate of a robot manipulation task.

\begin{figure}[t]
    \centering
    \includegraphics[width=\columnwidth]{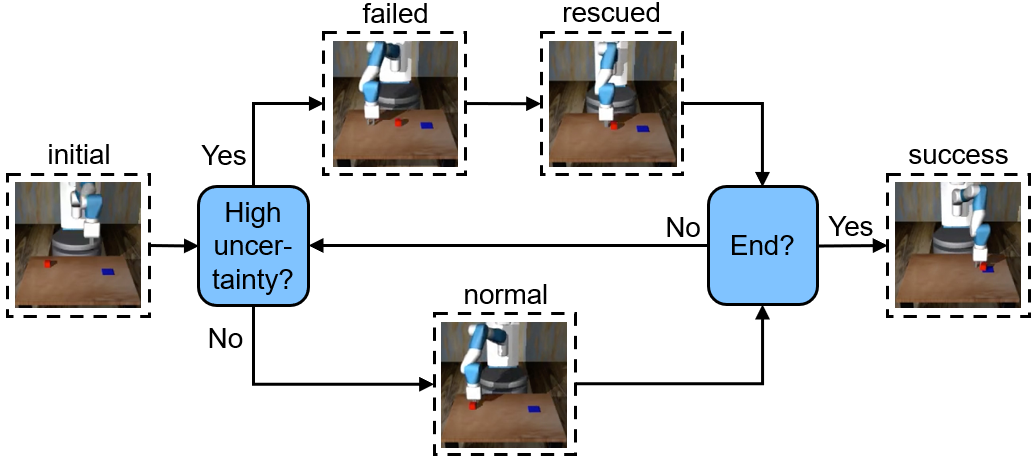}
    \caption{An overview of the proposed VMC approach with failure case recovery. In this example, the task is to push the red cube onto the target.}
    \label{fig:overall}
\end{figure}
\section{RELATED WORK}
The problem we are considering is based on learning robot control from visual feedback and monitoring policy uncertainty to optimise overall task success rate.
Our solution builds upon visuomotor control, uncertainty estimation and failure case recovery.

\textbf{{Visuomotor Control.}}
To plan robot motion from visual feedback, an established line of research is to use visual model-predictive control. The idea is to learn a forward model of the world, which forecasts the outcome of an action.  
In the case of robot control, a popular approach is to learn the state-action transition models in a latent feature embedding space, which are further used for motion planning~\cite{watter2015embed, agrawal2016learning, yu2019unsupervised}.
Likewise, visual foresight~\cite{finn2017deep} leverages a deep video prediction model to plan the end-effector motion by sampling actions leading to a state which approximates the goal image.
However, visual model-predictive control relies on learning a good forward model, and sampling suitable actions is not only computationally expensive but also requires finding a good action distribution.
End-to-end methods solve the issues mentioned above by directly predicting the next action.
Guided policy search~\cite{levine2016end} was one of the first to employ an end-to-end trained neural network to learn visuomotor skills, yet their approach requires months of training and multiple robots.
Well-known imitation learning approaches such as GAIL~\cite{ho2016generative} and SQIL~\cite{reddy2019sqil}
could also serve as backbones upon which we build our probabilistic approach.
However, we chose end-to-end visuomotor control~\cite{james2017transferring} as our backbone network architecture, for its simplicity and ability to achieve a zero-shot sim-to-real adaption through domain randomisation.

\textbf{Uncertainty Estimation.}
Approaches that can capture predictive uncertainties such as Bayesian Neural Networks~\cite{mackay1992practical} and Gaussian Processes~\cite{rasmussen2003gaussian} usually lack scalability to big data due to the computational cost of inferring the exact posterior distribution.
Deep neural networks with dropout~\cite{srivastava2014dropout} address this problem by leveraging variational inference~\cite{jordan1999introduction} and imposing a Bernoulli distribution over the network parameters.
The dropout training can be cast as approximate Bayesian inference over the network’s weights~\cite{gal2016dropout}. 
Gal et al.\@~\cite{gal2015bayesian} show that for the deep convolutional networks with dropout applied to the convolutional kernels, the uncertainty can also be computed by performing Monte Carlo sampling at the test phase. 
Rather than doing a grid search over the dropout rate which is computationally expensive, concrete dropout~\cite{gal2017concrete} relaxes the discrete Bernoulli distribution to the concrete distribution and thus allows the dropout rate to be trained jointly with other model parameters using the reparameterisation trick~\cite{kingma2013auto}.


\textbf{Failure Case Recovery.}
Most of the existing research utilise the fast inference of deep models to achieve closed-loop control~\cite{zhu2018reinforcement, pinto2017asymmetric, matas2018sim}. However, failure case detection and recovery in continuous operation has not been considered in other works.
Moreover, predicted actions are usually modelled as deterministic~\cite{srinivas2018universal, lee2019follow}, while the uncertainty of the policy networks has not been thoroughly investigated.
Another line of research considering failure recovery is interactive imitation learning, which assumes access to an oracle policy.
Similar to our work, HG-DAgger~\cite{kelly2019hg} estimates the epistemic uncertainty in an imitation learning setting, but by formulating their policy as an ensemble of neural networks, and they use the uncertainty to determine at which degree a human should intervene.
In this paper, our intuition is to detect the failure cases by monitoring the uncertainty of the policy neural network and rescue the robot when it is likely to fail by exploring into the robot state-action space under high confidence (i.e. low uncertainties).

\begin{figure}[tbh]
    \centering
    \includegraphics[width=\columnwidth]{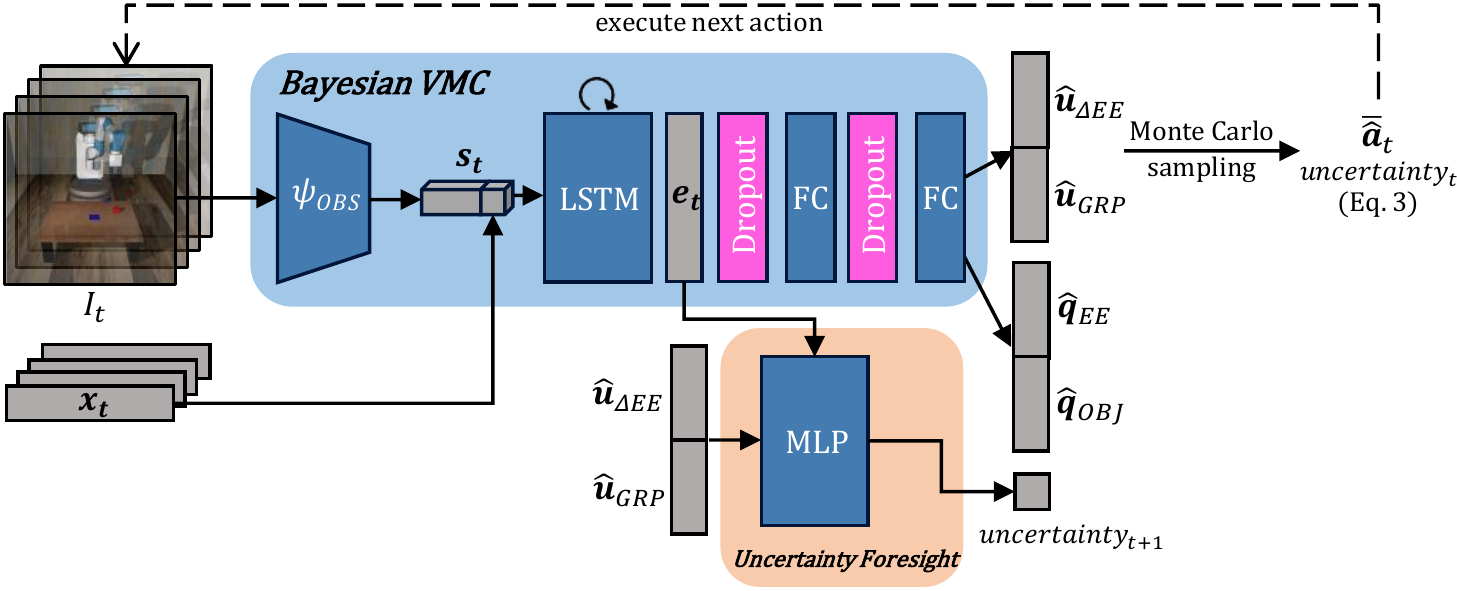}
    \caption{Network architecture of Introspective Visuomotor Control model. Blue: the backbone Bayesian Visuomotor Control model. The current observation $I_t$ is passed through a CNN $\psi_{OBS}$. This spatial feature map is concatenated to the tiled proprioceptive feature $\mathbf{x}_t$. The concatenated state representation $\mathbf{s}_t$ is fed into an LSTM. The LSTM embedding $\mathbf{e}_t$ is passed through a number of concrete dropout layers and fully connected layers interleavingly, whose output is then decoded into action commands $\mathbf{\hat{u}}_{\Delta EE}$ and $\mathbf{\hat{u}}_{GRP}$ as well as auxiliary position predictions $\mathbf{\hat{q}}_{EE}$ and $\mathbf{\hat{q}}_{OBJ}$.
    During test time, the mean action $\mathbf{\overline{\hat{a}}}_t$ is executed as the next action. The uncertainty estimate of the next timestep is used to supervise the prediction of the uncertainty foresight model.
    Orange: uncertainty foresight model. The LSTM embedding $\mathbf{e}_t$ is concatenated with the action commands $\mathbf{\hat{u}}_{\Delta EE}$ and $\mathbf{\hat{u}}_{GRP}$. It is passed through an MLP with 2 fully connected layers to predict the uncertainty associated with the next embedding $\mathbf{e}_{t+1}$.} 
    \label{fig:method-ivmc-model}
\end{figure}

\section{MODELLING UNCERTAINTY IN DEEP VISUOMOTOR CONTROL}


To detect the potential failure cases in manipulation, we build a probabilistic policy network for VMC.
Uncertainty is viewed as an indicator of the likelihood of task failure.

\textbf{End-to-End Visuomotor Control.}
For clarity, we first briefly review the end-to-end visuomotor control model~\cite{james2017transferring}. At timestep $t$, it takes $K$ consecutive frames of raw RGB images $(I_{t-K+1}, ..., I_t)$ as input to a deep convolutional neural network and outputs the embedding $(\mathbf{e}_{t-K+1}, ..., \mathbf{e}_t)$. To incorporate the configuration space information, the embedding is first concatenated with the corresponding robot joint angles $(\mathbf{x}_{t-K+1}, ..., \mathbf{x}_t)$ and then fed into a recurrent network followed by a fully connected layer. The buffered history information of length $K$ is leveraged to capture the higher-order states, e.g. the velocity and acceleration. In an object manipulation task using a robot gripper, the model predicts the next joint velocity command $\mathbf{\hat{u}}_J$ and the next discrete gripper action $\mathbf{\hat{u}}_{GRP}$ (open, close or no-op) as well as the object position $\mathbf{\hat{q}}_{OBJ}$ and gripper position $\mathbf{\hat{q}}_{EE}$ as auxiliary targets with the following loss objective:
%
%
\begin{equation}
\begin{aligned}
    \mathcal{L}_{\operatorname{total}} = & \operatorname{MSE}(\mathbf{\hat{u}}_J, \mathbf{u}_J)
    + \operatorname{CCE}(\mathbf{\hat{u}}_{GRP}, \mathbf{u}_{GRP}) \\
    & + \operatorname{MSE}(\mathbf{\hat{q}}_{OBJ}, \mathbf{q}_{OBJ})
    + \operatorname{MSE}(\mathbf{\hat{q}}_{EE}, \mathbf{q}_{EE}),
\end{aligned}
\label{eq:loss-vmc}
\end{equation}
%
%
%
where $\operatorname{MSE}$ and $\operatorname{CCE}$ stand for \emph{Mean-Squared Error} and \emph{Categorical Cross-Entropy} respectively. The losses are equally weighted and the model is trained end-to-end with stochastic gradient descent.

In this work, we use delta end-effector position command $\mathbf{\hat{u}}_{\Delta EE}$ rather than joint velocity command $\mathbf{\hat{u}}_J$ as a model output. We have found this to be more stable and less prone to the accumulated error over a long time horizon.
We feed a buffer of $K=4$ input frames at every timestep, and as we rollout the model, we keep the LSTM memory updated along the whole trajectory, as opposed to just $K$ buffered frames.

\textbf{Uncertainty Estimation.}
\label{subsection:method-estimating-uncertainty}
In the Bayesian setting, the exact posterior distribution of the network weights is intractable in general, due to the marginal likelihood.
In the variational inference case, we consider an approximating variational distribution, which is easy to evaluate.
To approximate the posterior distribution, we minimise the Kullback-Leibler divergence between the variational distribution and the posterior distribution.
Gal et al.~\cite{gal2016dropout} propose using dropout as a simple stochastic regularisation technique to approximate the variational distribution.
Training a deep visuomotor control policy with dropout not only reduces overfitting, but also enforces the weights to be learned as a distribution and thus can be exploited to model the epistemic uncertainty.

In practice, we train a Bayesian dropout visuomotor control policy and
evaluate the posterior action command distribution by integrating Monte Carlo samples.
At test time, we rollout the policy by performing stochastic forward passes at each timestep.
Figure~\ref{fig:method-ivmc-model} depicts the network architecture of our model.
To learn the dropout rate adaptively, we add concrete dropout layers.
Concrete dropout~\cite{gal2017concrete} uses a continuous relaxation of dropout's discrete masks and enables us to train the dropout rate as part of the optimisation objective, for the benefit of providing a well-calibrated uncertainty estimate.
We also experiment with the number of dropout layers.
We choose one and two layers since we do not want to add unnecessary trainable parameters and increase the computation cost.
The number of fully connected layers is adjusted according to that of dropout layers.

At timestep $t$, we draw
action samples $A_t=\{\mathbf{\hat{a}}_t^1, \mathbf{\hat{a}}_t^2, ...\}$,
where $\mathbf{\hat{a}}_t^i = [\mathbf{\hat{u}}_{\Delta EE,t}^i, \mathbf{\hat{u}}_{GRP,t}^i]^T$ is a model output, and use their mean $\mathbf{\overline{\hat{a}}}_t=\operatorname{mean}(A_t)$ as the action command to execute in the next iteration.
For an uncertainty estimate, following probabilistic PoseNet~\cite{kendall2016modelling}, we have experimented with the trace of covariance matrix of the samples and the maximum of the variance along each axis.
Similarly, we have found the trace to be a representative scalar measure of uncertainty. 

Simply computing the trace from a batch of sampled action commands does not capture the uncertainty accurately in cases where the predicted values vary significantly in norm in an episode.
For instance, when the end-effector approaches an object to interact with, it needs to slow down. At such a timestep, since the predicted end-effector commands are small, the trace of the covariance matrix is also small.
To calibrate the uncertainty measure, we transform every predicted delta end-effector position command $\mathbf{\hat{u}}_{\Delta EE}$ into norm and unit vector, weight them with $\lambda$ and $1-\lambda$ respectively, and concatenate them as a 4-dimensional vector $\mathbf{\hat{X}}$, before computing the trace:
\begin{equation}
\begin{aligned}
    & \mathbf{\hat{u}}_{\Delta EE} = [\hat{u}_x, \hat{u}_y, \hat{u}_z]^T \mapsto \mathbf{\hat{X}} = \\
    & 
    \mbox{$\squeezespaces{0.5}
    [\lambda \left \| \mathbf{\hat{u}}_{\Delta EE}  \right \|, (1-\lambda) \frac{\hat{u}_x}{\left \| \mathbf{\hat{u}}_{\Delta EE} \right \|}, (1-\lambda) \frac{\hat{u}_y}{\left \| \mathbf{\hat{u}}_{\Delta EE} \right \|}, (1-\lambda) \frac{\hat{u}_z}{\left \| \mathbf{\hat{u}}_{\Delta EE} \right \|}]^T.$}
\end{aligned}
\label{eq:weighted-concat}
\end{equation}
Here $\lambda$ is treated as a hyper-parameter. The superscripts $i$ denoting sample id and the subscripts $t$ denoting timestep are omitted for readability.

To determine how many Monte Carlo samples are required to achieve convergence, we compare the predicted action commands with the ground truth in validation episodes.
We compute the median error in each episode and average over validation episodes.
Monte Carlo sampling converges after around 50 samples and no more improvement is observed with more samples.
We thus define:

\begin{equation}
    \operatorname{uncertainty}_t = \operatorname{Tr}\Big(\operatorname{cov}\big([\mathbf{\hat{X}}_t^1, \mathbf{\hat{X}}_t^2, ..., \mathbf{\hat{X}}_t^{50}]^T\big)\Big),
\label{eq:unc-tr-cov}
\end{equation}
where $\mathbf{\hat{X}}_t^{i} \in \mathbb{R}^{4\times 1}$ is a sampled prediction transformed into weighted norm and unit vector in Eq.~\ref{eq:weighted-concat}.



\section{RECOVERY FROM FAILURES}

Our Bayesian visuomotor control model provides us with an uncertainty estimate of the current state at each timestep. In this section, we describe how we make use of it to recover from failures. 

\textbf{Knowing When to Recover.}
Continuously executing an uncertain trajectory is likely to lead to failure; diagnosis in an early stage and recovery can bring execution back on track. The question is, at which point shall we switch to a recovery mode to optimise overall success? Having a Bayesian VMC model trained, we deploy it on validation episodes to pick an optimal threshold of uncertainty for recovery.
Section~\ref{subsection:exp-uncertainty-threshold} details how to pick this threshold.  
During test time, as we rollout the model, when the uncertainty estimate is over the threshold, we switch to a recovery mode.


\textbf{Following Minimum Uncertainty.}
\label{subsection:method-follow-min-uncertainty}
Once the robot is switched to a recovery mode, our intuition is to explore in the state-action space and modify the robot configuration to an area trained with sufficient training examples. Hence, we propose moving along the trajectory with minimisation of uncertainty. However, the uncertainty estimate from the Bayesian VMC model in Figure~\ref{fig:method-ivmc-model} is associated with the current state. The Bayesian VMC model cannot provide the uncertainty of future frames without physically trying it. 
To address this issue, drawing inspiration from Embed to Control~\cite{watter2015embed} which
extracts a latent dynamics model for control from raw images,
we came up with the idea of learning a transition model mapping from the current latent feature embedding $\mathbf{e}_t$ given by our Bayesian VMC model to future $\mathbf{e}_{t+1}$ conditioned on an action $\mathbf{a}_t$. 
Then the predicted feature embedding $\mathbf{e}_{t+1}$ could be
fed as input to the first dropout layer through the last fully connected layer
to sample actions and estimate the uncertainty.
However, this approach of predicting next embedding $\mathbf{e}_{t+1}$ conditioned on action $\mathbf{a}_t$ would require further Monte Carlo sampling to estimate the uncertainty, making it computationally costly during test time.

Instead of predicting in the latent space, inspired by Visual Foresight~\cite{finn2017deep}, we predict the uncertainty of the next embedding $\mathbf{e}_{t+1}$ after executing $\mathbf{a}_t$ directly.
This can be achieved by Knowledge Distillation~\cite{bulo2016dropout}.
Specifically, we use the model uncertainty of time t+1 as the learning target to train the uncertainty foresight model.
We refer the reader to Figure~\ref{fig:method-ivmc-model}.

During test time, when the minimum uncertainty recovery mode is activated, we first backtrack the position of the end-effector to a point of minimum uncertainty within 20 steps. This is implemented by storing action, LSTM memory, uncertainty estimate and timestep in a FIFO queue of a maximum size of 20. Although the original state cannot always be recovered exactly in the case when the object is moved or when considering sensing and motor noise on a real system, backtracking guides the robot back into the vicinity of states where previous policy execution was confident. Then, at each timestep, we sample actions from the Bayesian VMC model and choose the action leading to the next state with minimum uncertainty according to our uncertainty foresight model. Algorithm~\ref{algo:method-failure-recovery} explains how this works within the Bayesian VMC prediction loop. With the same minimum recovery interval, we have observed that it is common to get stuck in a recovery loop, where after recovery the robot becomes too uncertain at the same place and goes into recovery mode again. Inspired by the binary exponential backoff algorithm -- an algorithm used to space out repeated retransmissions of the same block of data to avoid network congestion -- we double the minimum recovery interval every time that the recovery mode is activated. This simple intuitive trick solves the problem mentioned above well empirically.


\begin{algorithm}[tb]
\caption{Failure recovery for Bayesian VMC (test time)}
\begin{algorithmic}[1]
\REQUIRE $f$: trained Bayesian VMC model, $g$: trained Bayesian VMC model and uncertainty foresight module, outputting the action with the minimum epistemic uncertainty among samples from $f$, $T_{recovery}$: minimum recovery interval, $S$: number of samples used to compute uncertainty, $C$: recovery threshold.

\STATE \textit{\# Rollout a trained model.} 

\WHILE{\TRUE}
\STATE Sample $S$ actions from $f$ and compute their mean and uncertainty estimate.
\STATE Update the sum of a sliding window of uncertainties.

\STATE \textit{\# Check if failure recovery is needed.}
\IF{time since last recovery attempt $ > T_{recovery}$ \AND uncertainty sum $> C$ }
\STATE \textit{\# Uncertainty is high: start recovery.}
\STATE Double $T_{recovery}$.
\STATE Update last recovery attempt timestep.
\STATE Backtrack to a position with min uncertainty within the last few steps; restore memory.
\STATE Rollout $g$ for a number of steps.
\ELSE
\STATE \textit{\# Uncertainty is low: perform a normal action.}
\STATE Execute the mean action command of Monte Carlo sampling from $f$.
\ENDIF

\IF{maximum episode steps reached \OR task success}
\STATE \textbf{break}
\ENDIF

\ENDWHILE

\STATE \RETURN binary task success

\end{algorithmic}
\label{algo:method-failure-recovery}
\end{algorithm}


\section{EXPERIMENTS}
\label{section:experiments}
Our experiments are designed to answer the following questions: \textbf{(1)} Is uncertainty computed from stochastic sampling from our Bayesian VMC models a good indication of how well the model performs in an episode? \textbf{(2)} How well can our model recover from failures? \textbf{(3)} How well does our proposed minimum uncertainty recovery strategy perform compared to other recovery modes?

\begin{figure}[tbh]
    \centering
    \includegraphics[width=\columnwidth]{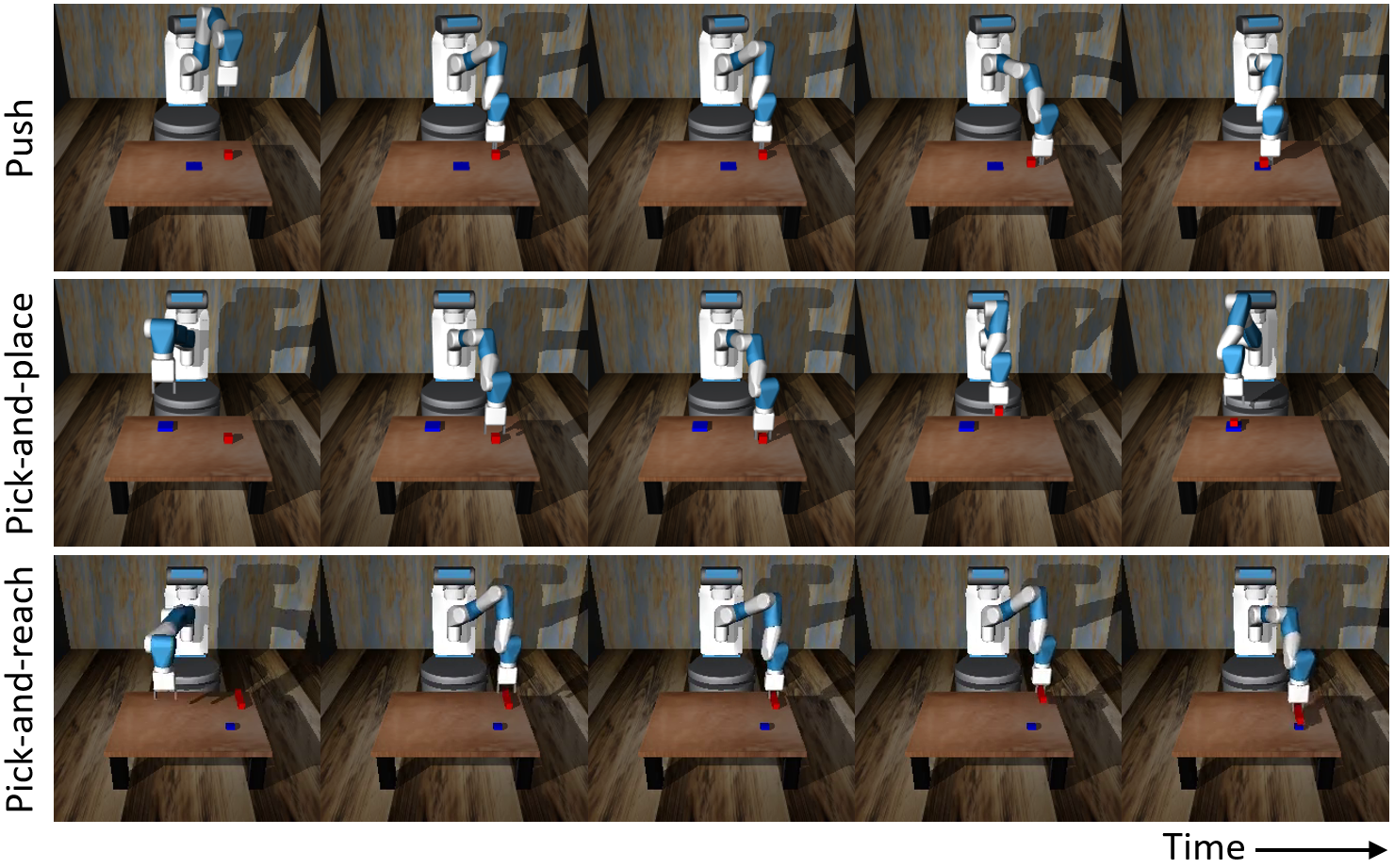}
    \caption{Top: Example of a pushing expert demonstration. The robot first pushes the red cube forward to align it with the blue target, and then moves to the side to push it sideways onto the target. Middle: Example of pick-and-place expert demonstration. The robot first moves toward the red cube to pick it up, and then moves to the blue target to drop the cube. Bottom: Example of a pick-and-reach expert demonstration. The robot first moves towards the red stick to pick it up at one end, and then reaches the blue target with the other end.}
    \label{fig:exp-push-demo}
\end{figure}

\textbf{Experimental Setup and Data Collection.}
We follow Gorth et al.\@~\cite{groth2020goal} and use the MuJoCo physics engine~\cite{todorov2012mujoco} along with an adapted Gym environment~\cite{openai2016gym} provided by~\cite{duan2017one} featuring the \emph{Fetch Mobile Manipulator}~\cite{wise2018fetch} with a 7-DoF arm and a 2-finger gripper.
Three tasks (Figure~\ref{fig:exp-push-demo}) are designed as they are fundamental in manipulation and commonly used as building blocks for more complex tasks.
In the pushing and pick-and-place tasks, the cube and the target are randomly spawned in a 6x8 grid, as opposed to only 16 initial cube positions and 2 initial target positions in the VMC~\cite{james2017transferring} pick-and-place task. In the pick-and-reach task, the stick and the target are spawned in 2 non-overlapping 6x8 grids. Similarly, we generate expert trajectories by placing pre-defined waypoints and solving the inverse kinematics. For each task, 4,000 expert demonstrations in simulation are collected, each
lasting 4 seconds long.
These are recorded as a list of observation-action tuples at 25 Hz, resulting in an episode length of $H = 100$.
For the uncertainty foresight model, we collect 2,000 trajectories from deploying a trained Bayesian VMC.
At every timestep, we execute an action sampled from the Bayesian VMC.
We record the current embedding, the action executed and the uncertainty of the next state after the action is executed, as described in Section~\ref{subsection:method-estimating-uncertainty}. 
An episode terminates after the task is completed or after the maximum episode limit of 200 is reached.

\textbf{Picking Uncertainty Threshold.}
\label{subsection:exp-uncertainty-threshold}
Uncertainty estimates can sometimes be noisy, so we smooth them out using a sliding window, given the assumption that uncertainties contiguously change throughout the course of a trajectory.
We have found a sliding window of 20 frames best avoids noisy peaks.
It is worth mentioning that the simulator runs at 25 Hz and 20 frames correspond to only 0.8 seconds.
For each evaluation episode, we record a binary label (i.e. task fail/success) and the maximum sum of a sliding window of uncertainties along the episode.
In the following, we denote the maximum sum of a sliding window of uncertainties as $u$ or maximum uncertainty.
We sort the episodes by their maximum uncertainty in increasing order.
Under the assumption that the probability of success after recovery is the overall average task success rate which is already known,
we pick a threshold to maximise the overall task success rate after recovery, which is equivalent to maximising the increase of successes.
We find the sorted episode index as follows.
\begin{equation}
\begin{aligned}
i^* = \underset{i}{\operatorname{argmax}}(
      & \left | \left \{ x \mid u(x)>u_{i} \right \} \right |\cdot \overline{r} \\
      & -\left | \left \{ x \mid u(x)>u_{i},\operatorname{result}(x)=\operatorname{success}) \right \} \right |),
\end{aligned}
\end{equation}
where $x$ is an episode, $u(x)$ is the maximum uncertainty of episode $x$, $u_i$ is the maximum uncertainty of episode indexed $i$, and $\overline{r}$ is the overall average success rate.



During test time, as we rollout the model, when the sum of a sliding window of 20 previous uncertainties is greater than the threshold of maximum uncertainty $u_{i^*}$, we switch to the recovery mode.

\textbf{Baselines for Visuomotor Control Recovery.}
\label{subsection:exp-recovery-baselines}
Our aim is to show our proposed failure recovery mode outperforms other failure recovery modes, as well the backbone VMC~\cite{james2017transferring}. Thus, we do not directly compare it against other visuomotor control approaches. 
We compare our failure recovery mode \begin{sc}min unc\end{sc} in Section~\ref{subsection:method-follow-min-uncertainty} against two baselines: \begin{sc}rand\end{sc} and \begin{sc}init\end{sc}.
The recoveries all happen when the uncertainty is high while deploying a Bayesian VMC (line 7 of Algorithm~\ref{algo:method-failure-recovery}).
We use a maximum of 25 recovery steps in all cases.
\textbf{(1)} \begin{sc}rand\end{sc}: The end-effector randomly moves 25 steps and we keep the gripper open amount as it is (no-op). Then, we reset the LSTM memory. \textbf{(2)} \begin{sc}init\end{sc}: We open the gripper, sample a point in a sphere above the table and move the end-effector to that point. Then, we reset the LSTM memory. This recovery mode is designed to reset to a random initial position.
All the recovery modes attempt to move the robot from an uncertain state to a different one, with the hope of it being able to interpolate from the training dataset starting from a new state. 

\section{RESULTS}

\begin{figure}[tbh]
    \centering
    \includegraphics[width=0.49\columnwidth]{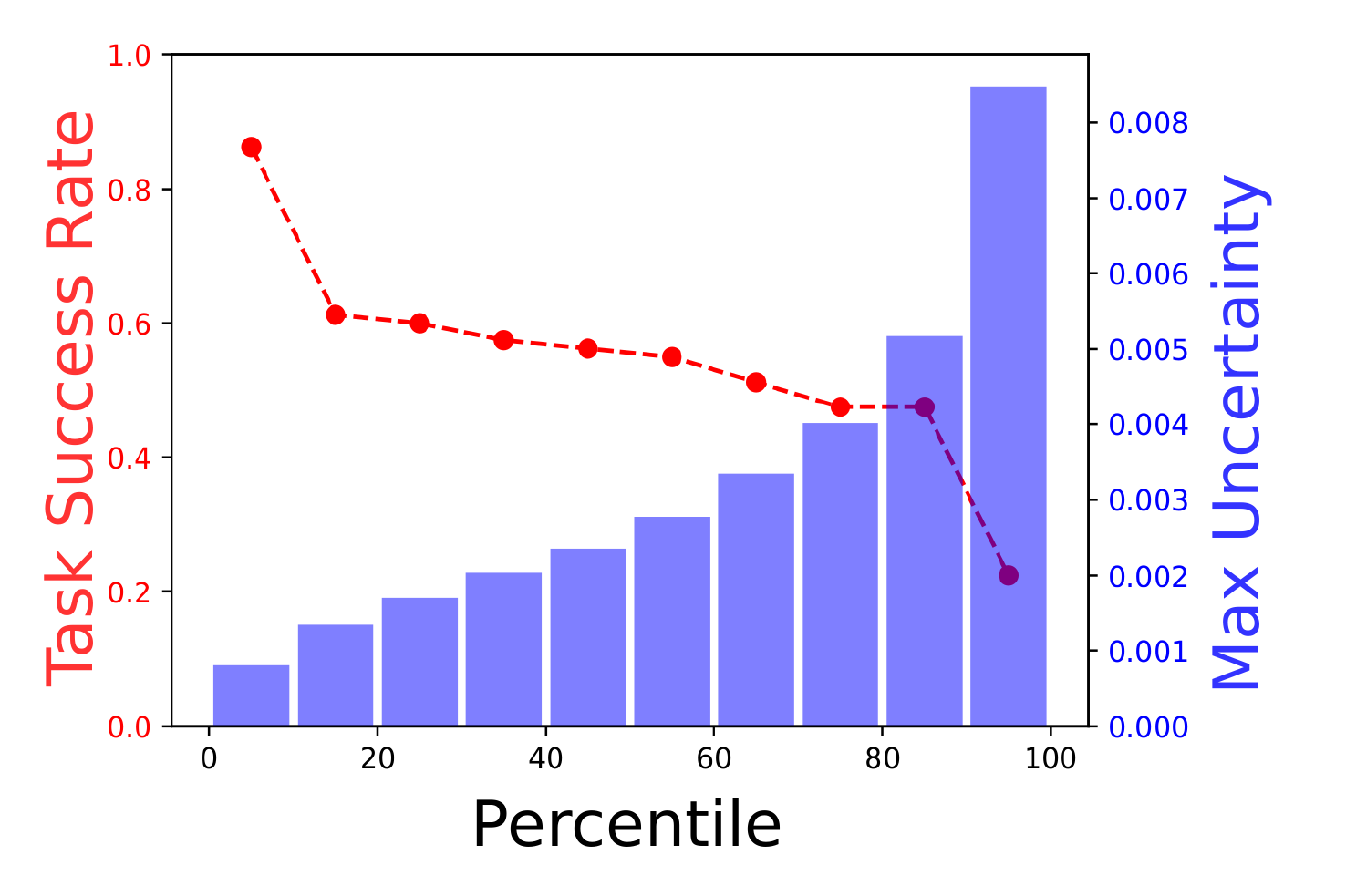}
    \includegraphics[width=0.49\columnwidth]{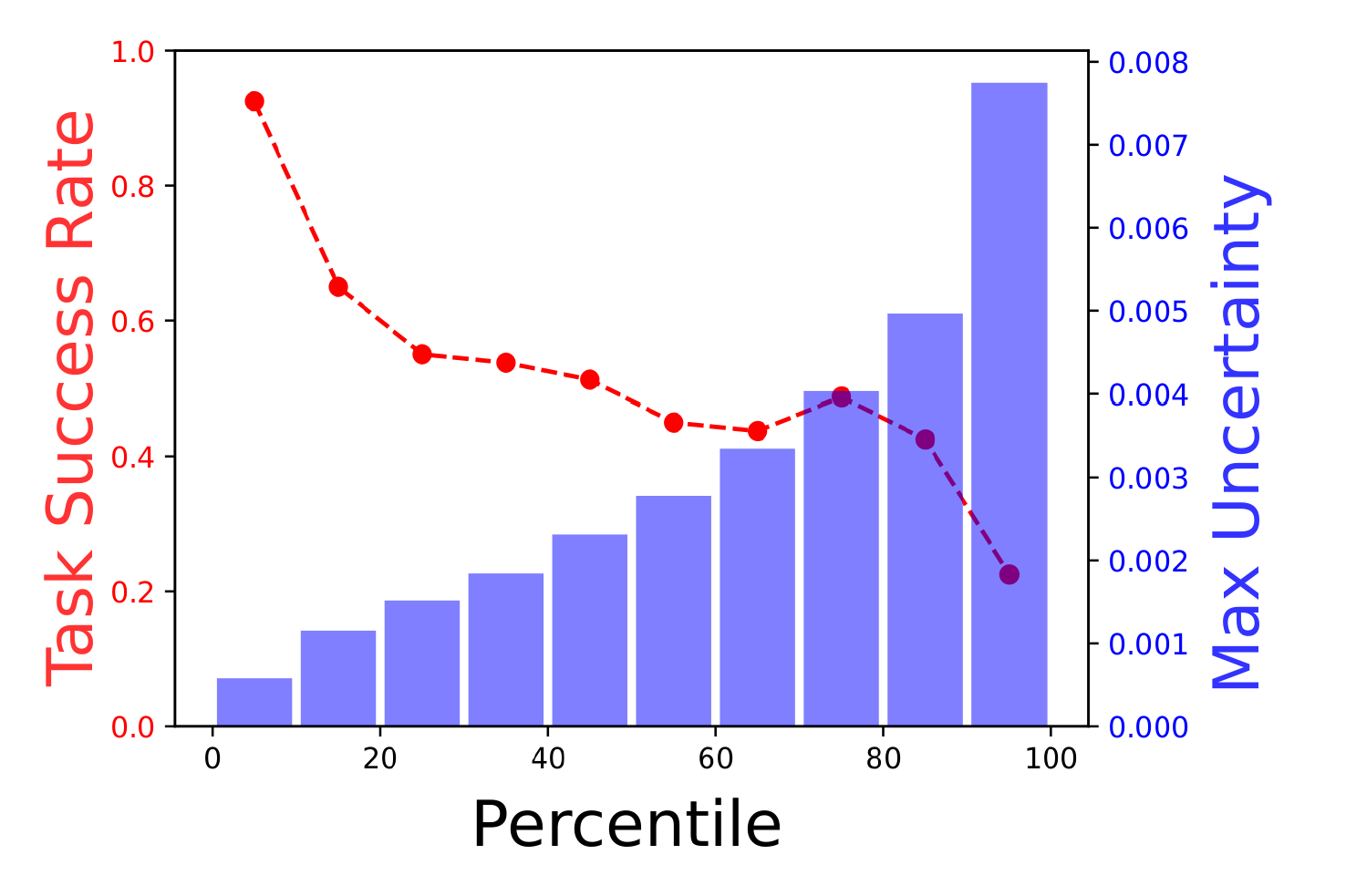}
    \includegraphics[width=0.49\columnwidth]{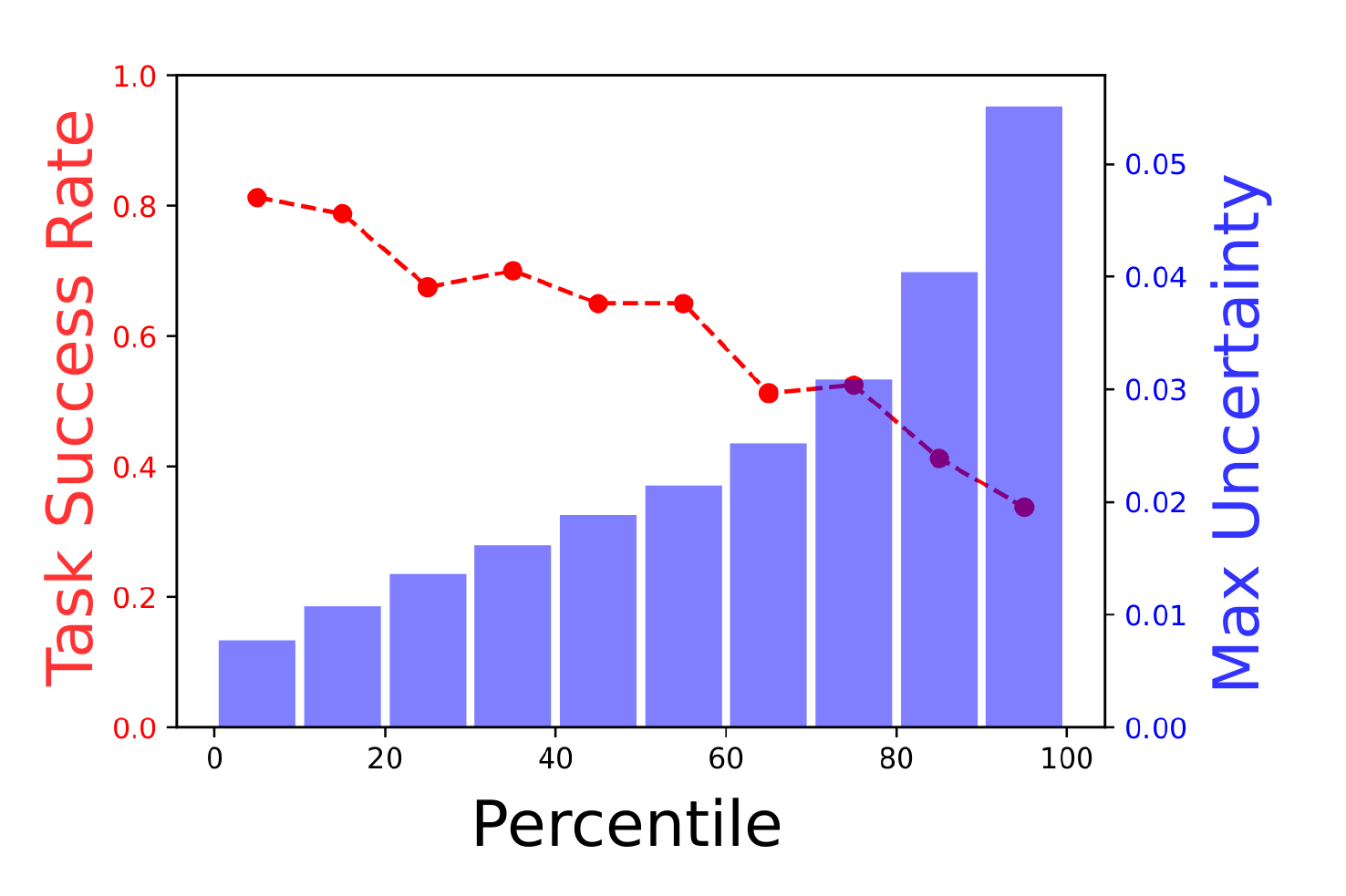}
    \includegraphics[width=0.49\columnwidth]{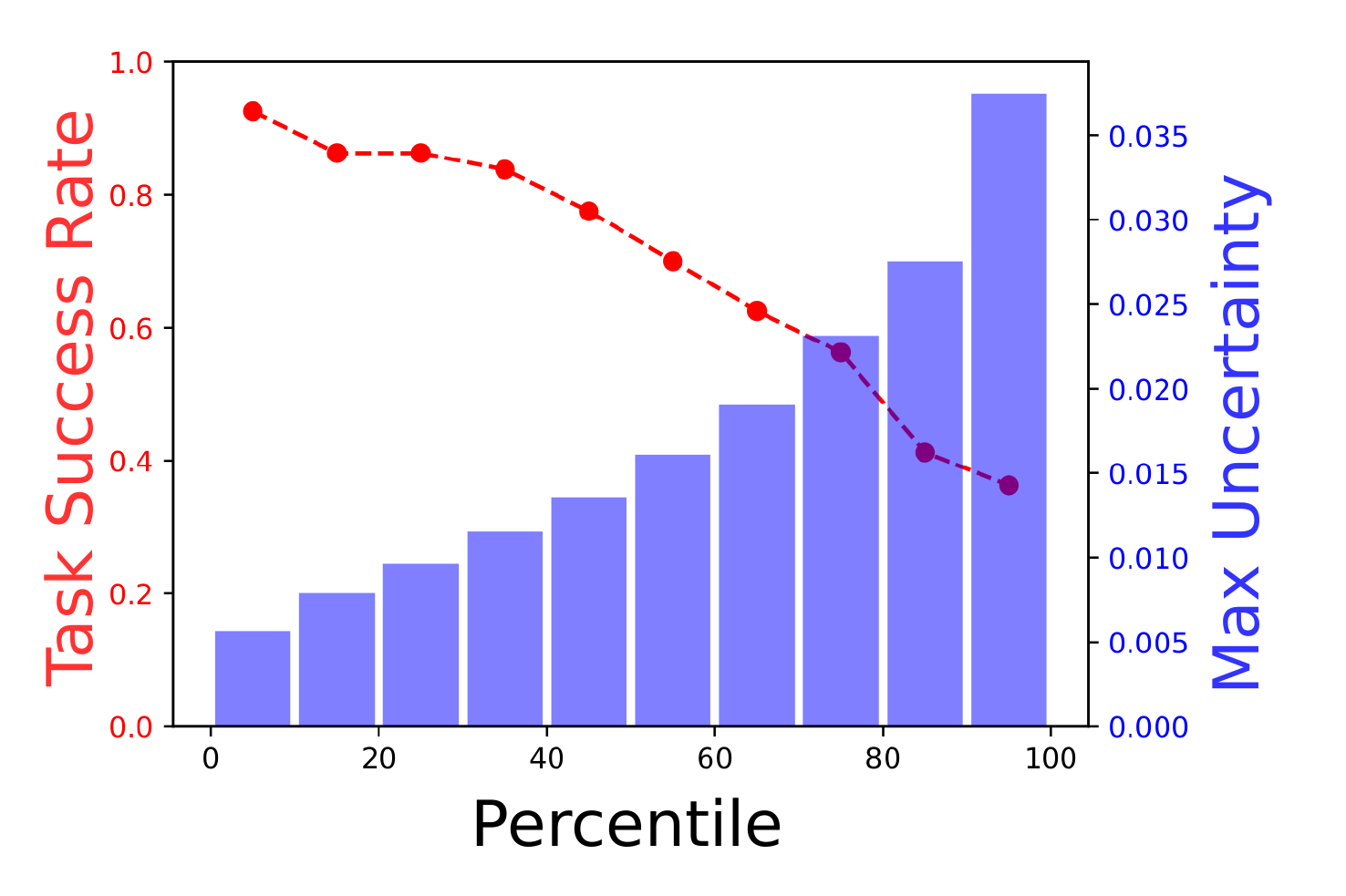}
    \caption{Evaluation of task success rate vs maximum uncertainty of different models evaluated over 800 test episodes. Left: one dropout layer. Right: two dropout layers. Top: pushing. Bottom: pick-and-place. These plots are drawn by sorting episodes by their maximum uncertainty and regrouping them into 10 bins. Subsequently, the average task success rate and the average maximum uncertainty are computed for each bin.}
    \label{fig:results-eval-concrete}
\end{figure}

\textbf{Task Success vs Uncertainty Estimate.}
Is uncertainty estimate a good indication of how well the model performs in an episode? To address this first guiding question in Section~\ref{section:experiments},
we analyse how the task success rate varies with respect to the uncertainty estimate from our Bayesian VMC models.
We evaluate on 800 test scene setups and regroup them by maximum uncertainty into 10 bins.
Figure~\ref{fig:results-eval-concrete} shows the task success rate versus maximum uncertainty in each bin.
We observe that task success rate is inversely correlated with maximum uncertainty, which corroborates our hypothesis of high uncertainty being more likely to lead to failure.

\textbf{Manipulation with Failure Recovery Results.}
Regarding the last two guiding questions in Section~\ref{section:experiments}, we evaluate the performance of the controllers on 100 held-out test scene setups for all three tasks.
We report all model performances in Table~\ref{table:result-push-pick-concrete}.

\begin{table*}[tbh]
\centering
\vskip 0.15in
\begin{center}
\begin{small}
\fontsize{7.585}{7.585}\selectfont
\begin{sc}
\begin{tabular}{l|cc|ccc|ccc}
\toprule
\multirow{3}{*}{model \#fc=1} & \multicolumn{2}{c|}{Pushing} & \multicolumn{3}{c|}{Pick-and-Place} & \multicolumn{3}{c}{Pick-and-Reach} \\
& \multicolumn{1}{c}{reach} & \multicolumn{1}{c|}{push} & \multicolumn{1}{c}{reach} & \multicolumn{1}{c}{pick} & \multicolumn{1}{c|}{place} & \multicolumn{1}{c}{reach} & \multicolumn{1}{c}{pick} & \multicolumn{1}{c}{task} \\
& {[}\%{]} & {[}\%{]} & {[}\%{]} & {[}\%{]} & {[}\%{]} & {[}\%{]} & {[}\%{]} & {[}\%{]} \\
\hline
VMC~\cite{james2017transferring} & \textbf{97.00 $\pm$ 1.62} & 49.00 $\pm$ 4.74 & 99.00 $\pm$ 0.94 & 77.00 $\pm$ 3.99 & 52.00 $\pm$ 4.74 & \textbf{99.00 $\pm$ 0.94} & 77.00 $\pm$ 3.99 & 69.00 $\pm$ 4.39 \\
\hline
BVMC & 91.00 $\pm$ 2.71 & 50.00 $\pm$ 4.75 & 99.00 $\pm$ 0.94 & 84.00 $\pm$ 3.48 & 60.00 $\pm$ 4.65 & \textbf{99.00 $\pm$ 0.94} & 88.00 $\pm$ 3.08 & 78.00 $\pm$ 3.93 \\
+ rand & 93.00 $\pm$ 2.42 & \textbf{56.00 $\pm$ 4.71} & 99.00 $\pm$ 0.94 & \textbf{85.00 $\pm$ 3.39} & 68.00 $\pm$ 4.43 & \textbf{99.00 $\pm$ 0.94} & \textbf{89.00 $\pm$ 2.97} & \textbf{81.00 $\pm$ 3.72} \\
+ init & 93.00 $\pm$ 2.42 & \textbf{55.00 $\pm$ 4.72} & 99.00 $\pm$ 0.94 & \textbf{88.00 $\pm$ 3.08} & 67.00 $\pm$ 4.46 & \textbf{99.00 $\pm$ 0.94} & \textbf{93.00 $\pm$ 2.42} & \textbf{79.00 $\pm$ 3.86} \\
+ min unc & \textbf{94.00 $\pm$ 2.25} & \textbf{58.00 $\pm$ 4.68} & 99.00 $\pm$ 0.94 & \textbf{90.00 $\pm$ 2.85} & 70.00 $\pm$ 4.35 & \textbf{99.00 $\pm$ 0.94} & \textbf{93.00 $\pm$ 2.42} & \textbf{82.00 $\pm$ 3.64} \\
\hline
\multirow{3}{*}{model \#fc=2} & \multicolumn{2}{c|}{Pushing} & \multicolumn{3}{c|}{Pick-and-Place}  & \multicolumn{3}{c}{Pick-and-Reach} \\
& \multicolumn{1}{c}{reach} & \multicolumn{1}{c|}{push} & \multicolumn{1}{c}{reach} & \multicolumn{1}{c}{pick} & \multicolumn{1}{c|}{place}  & \multicolumn{1}{c}{reach} & \multicolumn{1}{c}{pick} & \multicolumn{1}{c}{task} \\
& {[}\%{]} & {[}\%{]} & {[}\%{]} & {[}\%{]} & {[}\%{]} & {[}\%{]} & {[}\%{]} & {[}\%{]} \\
\hline
VMC~\cite{james2017transferring} & \textbf{96.00 $\pm$ 1.86} & 50.00 $\pm$ 4.75 & 97.00 $\pm$ 1.62 & 79.00 $\pm$ 3.86 & 60.00 $\pm$ 4.65 & \textbf{99.00 $\pm$ 0.94} & 79.00 $\pm$ 3.86 & 70.00 $\pm$ 3.64 \\
\hline
BVMC & 88.00 $\pm$ 3.08 & \textbf{53.00 $\pm$ 4.74} & \textbf{100.00 $\pm$ 0.00} & \textbf{87.00 $\pm$ 3.19} & 69.00 $\pm$ 4.39 & \textbf{99.00 $\pm$ 0.94} & \textbf{89.00 $\pm$ 2.97} & \textbf{79.00 $\pm$ 3.86} \\
+ rand & 88.00 $\pm$ 3.08 & \textbf{60.00 $\pm$ 4.65} & \textbf{100.00 $\pm$ 0.00} & \textbf{91.00 $\pm$ 2.71} & 74.00 $\pm$ 4.16 & \textbf{99.00 $\pm$ 0.94} & \textbf{91.00 $\pm$ 2.71} & \textbf{82.00 $\pm$ 3.64} \\
+ init & 93.00 $\pm$ 2.42 & \textbf{58.00 $\pm$ 4.68} & \textbf{100.00 $\pm$ 0.00} & \textbf{89.00 $\pm$ 2.97} & \textbf{76.00 $\pm$ 4.05} & \textbf{99.00 $\pm$ 0.94} & \textbf{93.00 $\pm$ 2.42} & \textbf{83.00 $\pm$ 3.56} \\
+ min unc & 91.00 $\pm$ 2.71 & \textbf{62.00 $\pm$ 4.61} & \textbf{100.00 $\pm$ 0.00} & \textbf{89.00 $\pm$ 2.97} & \textbf{82.00 $\pm$ 3.64} & \textbf{99.00 $\pm$ 0.94} & \textbf{94.00 $\pm$ 2.25} & \textbf{85.00 $\pm$ 3.39} \\
\bottomrule
\end{tabular}

\end{sc}
\end{small}
\end{center}
\vskip 0.1in
\caption{
Comparison of model performances with and without failure recovery in the pushing, pick-and-place and pick-and-reach tasks.
Top: one fully connected layer. Bottom: two fully connected layers. Best task performances are bold-faced.
}
\label{table:result-push-pick-concrete}
\end{table*}


In the first row, we compare against \begin{sc}VMC\end{sc}, the original deterministic VMC model~\cite{james2017transferring}, but with one or two fully connected layers after the LSTM.
Next, \begin{sc}BVMC\end{sc}, the Bayesian VMC model executing the mean of the sampled predictions at each timestep, but not using the uncertainty estimate information for recovery.
Although this does not perform any recovery, the network architecture is slightly different than \begin{sc}VMC\end{sc} due to the added concrete dropout layer(s).
\begin{sc}BVMC + rand\end{sc} and \begin{sc}BVMC + init\end{sc} are the baseline recovery modes (Section~\ref{subsection:exp-recovery-baselines}).
Last, we present \begin{sc}BVMC + min unc\end{sc}, our proposed recovery mode following minimum uncertainty (Section ~\ref{subsection:method-follow-min-uncertainty}).

In the pushing task, although the reaching performance of \begin{sc}BVMC\end{sc} drops compared to \begin{sc}VMC\end{sc}, the pushing performance is slightly better. In general, adding stochasticity and weight regularisation prevents overfitting, but it does not always boost performance. \begin{sc}BVMC + rand\end{sc} and \begin{sc}BVMC + init\end{sc} outperform \begin{sc}BVMC\end{sc} by approximately 5\% in both cases of one and two fully connected layers.
The performance increase is moderate because a large proportion of bins of episodes in the mid maximum uncertain range has a task success rate close to the average overall task success rate
(Figure~\ref{fig:results-eval-concrete})
and the threshold of maximum uncertainty picked is relatively high, thus not allowing many episodes to switch to a recovery mode.
In general, the models with two fully connected layers have higher performance than their counterparts with one fully connected layer. This can be understood as having more trainable parameters helps learn a better function approximation. Our proposed \begin{sc}BVMC + min unc\end{sc} surpasses other two baseline recovery modes, indicating that following actions with minimum uncertainty contributes further to the task success.  

In pick-and-place and pick-and-reach, all VMC and Bayesian VMC models exhibit near perfect reaching performance. Also, surprisingly, all models do better than their counterparts in the pushing task. At first glance, both tasks seem to be more difficult than pushing. In fact, the design of our pushing task requires a two-stage rectangular push. We observe most failure cases in pushing happen when the end-effector does not push at the centre of the cube, so that the cube is pushed to an orientation never seen in the training dataset. This rarely happens in the pick-and-place and pick-and-reach tasks. Similarly, \begin{sc}BVMC + rand\end{sc} and \begin{sc}BVMC + init\end{sc} show a performance increase compared to \begin{sc}BVMC + no\end{sc}. Last but not least, \begin{sc}BVMC + min unc\end{sc} almost surpasses all other models in reaching, picking and placing/task, with a task success rate increase of 22\% compared to \begin{sc}VMC\end{sc} for pick-and-place and 15\% for pick-and-reach.

Qualitatively, we observe interesting behaviours from our uncertainty estimates and recovery modes. In all three tasks, when a Bayesian VMC controller approaches the cube with a deviation to the side, we often see the controller fall into the recovery mode, while a VMC controller with the same scene setup continues the task and eventually get stuck in a position without further movements. Occasionally, in the pick-and-place and pick-and-reach tasks when the end-effector moves up without grasping the cube successfully, the Bayesian VMC controller monitors high uncertainty and starts recovery.


\begin{figure}[bth]
    \centering
    \includegraphics[width=\columnwidth]{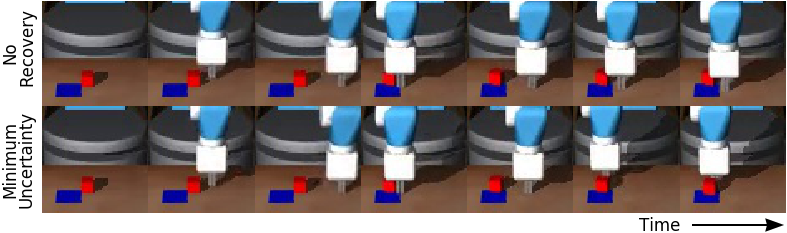}
    \caption{Recovery comparison. The top row depicts operation without recovery, while the bottom row shows the results with recovery based on the minimum uncertainty. The robot fails to accomplish the pushing task without the recovery. The images are cropped to emphasise the difference.
    }
    \label{fig:exp-push-compare}
\end{figure}







\textbf{System Efficiency.}
Recovery from uncertain states improves task performance. However, drawing stochastic samples also comes at an additional time cost.
By design of our network architecture, only the last dropout layers and fully connected layers need to be sampled, since the first 8 layers of convolutional neural network and LSTM are deterministic. 
For reference, on an NVIDIA GeForce GTX 1080, averaging 50 Monte Carlo samples and computing the uncertainty take around 0.1 seconds, while the original VMC takes around 0.03 seconds per timestep.
If treating the inference as a mini-batch of operations, this extra computation can be further reduced~\cite{gal2016uncertainty}.
\section{CONCLUSIONS}

This paper investigates the usage of policy uncertainty for failure case detection and recovery.
In our method, a Bayesian neural network with concrete dropout is employed to obtain the model epistemic uncertainty by Monte Carlo sampling.
We further make use of a deterministic model and knowledge distillation to learn the policy uncertainty of a future state conditioned on an end-effector action.
Consequently, we are able to predict the uncertainty of a future timestep without physically simulating the actions.
The experimental results verified our hypothesis -- the uncertainties of the VMC policy network can be used to provide intuitive feedback to assess the failure/success in manipulation tasks, and, reverting and driving the robot to a configuration with minimum policy uncertainty can recover the robot from potential failure cases.

\addtolength{\textheight}{-6cm}   



\section*{ACKNOWLEDGMENT}
Chia-Man Hung is funded by the Clarendon Fund and receives a Keble College Sloane Robinson Scholarship at the University of Oxford.
Yizhe Wu is funded by the China Scholarship Council.
This research is also supported by an EPSRC Programme Grant [EP/M019918/1] and a gift from Amazon Web Services (AWS).
The authors acknowledge the use of the University of Oxford Advanced Research Computing (ARC) facility in carrying out this work (\texttt{http://dx.doi.org/10.5281/zenodo.22558}).
We also thank Ruoqi He, Hala Lamdouar, Walter Goodwin and Oliver Groth for proofreading and useful discussions, and the reviewers for valuable feedback.


\bibliographystyle{IEEEtran}
\bibliography{ivmc}

\end{document}